\def\year{2022}\relax
\documentclass[letterpaper]{article} 
\usepackage{aaai22}  
\usepackage{times}  
\usepackage{helvet}  
\usepackage{courier}  
\usepackage[hyphens]{url}  
\usepackage{graphicx} 
\usepackage{natbib}  
\usepackage{caption} 
\usepackage{multicol}
\DeclareCaptionStyle{ruled}{labelfont=normalfont,labelsep=colon,strut=off} 
\usepackage{pifont}
\newcommand{\cmark}{\ding{51}}%
\newcommand{\xmark}{\ding{55}}%
\usepackage{color}
\usepackage{makecell}

\title{Re-Examining Human Annotations for Interpretable NLP}
\author{
    Cheng-Han Chiang, Hung-yi Lee
}
\affiliations{
    National Taiwan University\\

    No. 1, Sec. 4, Roosevelt Rd., Taipei 10617, Taiwan (R.O.C.) \\
    dcml0714 at gmail.com, hungyilee at ntu.edu.tw
}

\usepackage{bibentry}

\begin{document}

\maketitle

\begin{abstract}
Explanation methods in Interpretable NLP often explain the model's decision by extracting evidence (rationale) from the input texts supporting the decision.
Benchmark datasets for rationales have been released to evaluate how good the rationale is. 
The ground truth rationales in these datasets are often human annotations obtained via crowd-sourced websites.
Valuable as these datasets are, the details on how those human annotations are obtained are often not clearly specified.
We conduct comprehensive controlled experiments using crowd-sourced websites on two widely used datasets in Interpretable NLP to understand how those unsaid details can affect the annotation results.
Specifically, we compare the annotation results obtained from recruiting workers satisfying different levels of qualification.
We also provide high-quality workers with different instructions for completing the same underlying tasks.
Our results reveal that the annotation quality is highly subject to the workers' qualification, and workers can be guided to provide certain annotations by the instructions.
We further show that specific explanation methods perform better when evaluated using the ground truth rationales obtained by particular instructions.
Based on these observations, we highlight the importance of providing complete details of the annotation process and call for careful interpretation of any experiment results obtained using those annotations.
\end{abstract}

\section{Introduction}
Deep learning models have stricken great success in miscellaneous tasks in natural language processing (NLP).
However, the black-box nature of these models makes users hesitate when applying them.
This makes Interpretable NLP to gain great attention.
An important topic in Interpretable NLP is explanations based on input features, which extracts snippets in the input text that support the model's prediction.
To quantify how useful those explanations are for humans, researchers collected ground truth explanations by asking workers to annotate important parts in the text using crowd-sourced website\citep{deyoung2020eraser, camburu2018snli, lehman2019inferring, rajani2019explain}.
By comparing the explanations from explanation methods against those explanations from the workers, we can quantify the \textit{plausibility}\footnote{The plausibility of an explanation method quantifies how useful the explanation is for human.} of an explanation method.
\begin{figure}[t!]
\centering
\includegraphics[clip, trim =  80 30 100 20, width=0.8\linewidth]{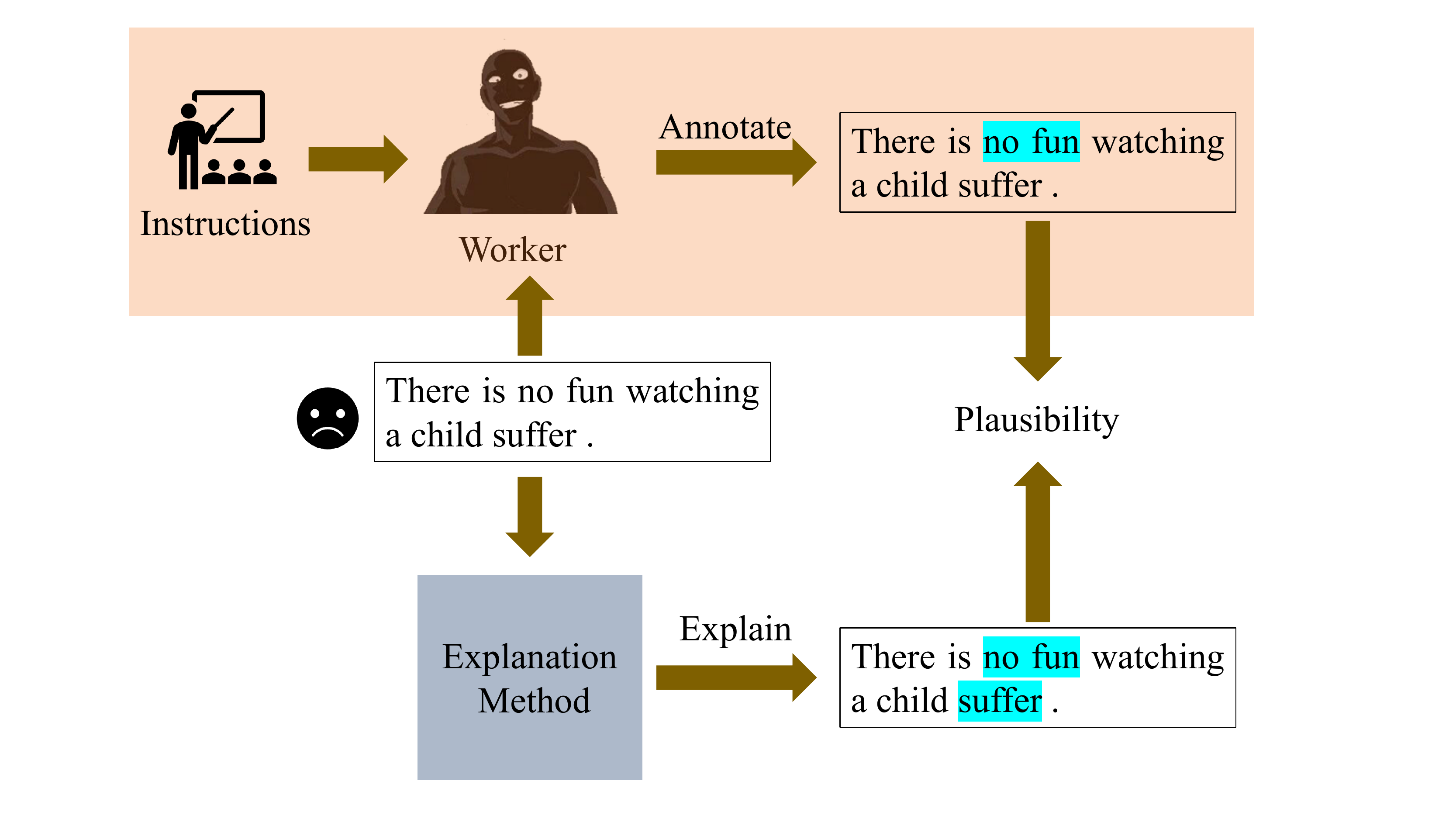}
\caption{The experiment setting in our work. 
For a piece of text and its ground truth answer, we ask the workers to follow the annotation instructions and label the important words in the text that support the label.
Meanwhile, we use some input feature-based explanation methods to generate the explanation.
By comparing the annotation result from humans and explanation by explanation method, we can quantify the plausibility of an explanation method.}
\label{fig:workflow}
\end{figure}

While these human annotations boost the researches in Interpretable NLP by enabling better comparison between different explanation methods, we identify that the detailed annotation processes in several important benchmark datasets are omitted. 
Some previous works did not specify how the workers for human annotation are selected, and others failed to report the precise instructions for the workers during the annotation process.
In the era that values reproducibility in terms of programming results, few have realized the importance of reproducibility of the human annotation process.

As a pioneering work, we aim to understand how those unspecified details during the annotation process can change the annotation results.
We focus on the following three research questions (RQs):
\textbf{RQ1}: How does the quality of the workers affect the annotation results?
 \textbf{RQ2}: How do the instructions of the annotation affect the annotation results? \textbf{RQ3}: Will annotations obtained from different instructions change the plausibility of a certain explanation method?

We construct systematic experiments and obtain the following findings:
\begin{itemize}
    \item We find that without placing any restrictions on the workers, the annotation results can be catastrophic. 
    Even when the workers fulfill some strict requirements (certificated by the crowd-sourcing website), the annotation results are not guaranteed to be good.
    \item We find that giving precise instructions will make the annotations more agreed among annotators for some datasets.
    Surprisingly, providing example annotations does not increase the agreement among annotators.
    We also show that annotators can be guided to provide annotations with a certain pattern.
    \item Some explanation methods' plausibility shows sensitivity to the instructions used for constructing the human annotations.  
\end{itemize}
To the best of our knowledge, we are the first to emphasize the importance of the details of the human annotation for Interpretable NLP by constructing controlled experiments.

\section{Preliminary}
\subsection{Interpreting Model Decision from Input Features}
\label{subsec:previous}
Explaining by input features aims to identify the parts in the input that are important for the model's decision.
The granularity of input features can range from words to sentences.
In this work, we restrict ourselves to considering the input features at the granularity of words.
Widely used input methods for scoring the word importance of the input texts include Gradient, Integrated Gradient (IG), Smooth Gradient (SG), Occlusion, and LIME\footnote{We provide a brief introduction of these input methods in Appendix~\ref{sec:input methods}.}

To evaluate how well the explanations provided by these input methods are aligned with humans, benchmark datasets containing explanations from humans are used.
In these human explanations, annotators are asked to label the parts in the input that support the ground truth answer, called rationales.
By comparing the agreement between the explanation provided by an input method and the annotation provided by the workers, we can quantify the plausibility of an explanation method.

However, in a widely used rationale benchmark datasets, ERASER ~\citep{deyoung2020eraser}, the annotation process of the seven sub-datasets is far from comprehensive.
Only three of the sub-dataset provide the information of the annotators; how the workers are instructed to annotate the rationales are unspecified\footnote{We refer to the test set rationales obtained by \citet{deyoung2020eraser}.}.
For the rest four datasets, little information about the workers can be found.
We review the annotation process of ERASER in Appendix~\ref{app:amt}.

\subsection{Online Crowd-Source Platforms}
Online crowd-sourced platforms are places where requesters can publish jobs and recruit workers from all around the world to perform their tasks.
We will briefly introduce Amazon Mechanical Turk (AMT), as it is the platform used in this paper.

In AMT, a worker is dubbed a turker.
Anyone who has access to the Internet and has an Amazon account can become a turker.
Every single task a turker works on is called a human intelligence task (HIT).
In our case, a single HIT is the rationale annotation of a text.
Turkers can be ranked by their HIT approval rate, which is the percentage of the HITs approved by the requesters among all the HITs a turker has submitted.
Higher HIT approval rates may indicate higher quality of the workforce.
AMT grants Master Qualifications to those who "consistently demonstrated a high degree of success in performing a wide range of HITs across a large number of requesters" based on some statistical models.

\section{Experiment Setting}
\label{sec:exp setting}
Our goal in this work is to understand how the selections of the workers and the instructions for the workers alter the annotation results.
To this end, we use AMT to annotate two commonly used datasets in Interpretable NLP while controlling the qualification of the turkers and varying the instructions for the turkers.

The two datasets we select are Common Sense Explanation (CoS-E) and SST-2. 
The former is adopted in ERASER~\citep{deyoung2020eraser} benchmark, and the latter can be seen as a shorter version of Movie Review~\citep{zaidan-etal-2007-using}.
For each dataset, we manually select 100 instances from the validation set or testing set as the annotation dataset.
We only select those instances with the correct text classification ground truth label\footnote{The ground truth labels of the two datasets are noisy.}, and we also only select those instances whose ground truth label can be easily predicted from the text without ambiguity.

During the annotation process, the turkers are provided with a piece of text and the ground truth answer in the dataset.
For each word in the text, the turkers are asked to determine whether the word is important for supporting the ground truth answer.
In CoS-E, we provide the turkers with the question, five options, and the ground truth answer.
Turkers need to select the important words from the question.
In SST-2, we provide the turkers with the movie review and the ground truth sentiment.
The annotators are required to select the important words from the movie review.
For both datasets, the turkers are required to select at least one important word in the text, otherwise, their annotation will be rejected.
We call the annotations that fail to label any important words the \textit{empty annotations}.
Each instance is annotated by three annotators.
The interface we use can be found in Figure~\ref{fig:instruction} and Figure~\ref{fig:interface}.

\section{How Do the Qualifications of Turkers Affect Annotated Results?}
\label{sec:qualification}
We first focus on how the qualifications of the turkers affect the annotation results.
We use the instructions in the first row of Table~\ref{tab:instruction}, and select turkers of different qualifications.
We use the HIT approval rate as an indicator of the quality of the workforce.
For each dataset, we recruit annotators with different HIT approval rates:  HIT approval rates higher than 80\%, 90\%, and 95\%. 
As a preliminary quality indicator of the workforce, we examine the average time a turker processes a HIT.
We expect that both tasks should take about 40 seconds for a turker to annotate an instance.

\begin{table}[h]
    \centering
    \begin{tabular}{c|c|c}
\hline
         HIT Approval Rate &CoS-E & SST-2  \\
         \hline
         $>$80\% & 22s & 71s \\ 
         $>$90\% & 30s & 27s\\
         $>$95\% & 52s & 30s\\
         \hline
    \end{tabular}
    \caption{Average time in seconds for a turker to annotate an instance.}
    \label{tab:worktime}
\end{table}
The results are shown in Table~\ref{tab:worktime}.
Observe that for CoS-E, turkers with higher HIT approval rates take more time annotating one instance on average.
Surprisingly, for SST-2, turkers with HIT approval rates larger than 80\% spent significantly longer on a single instance on average.
We find that about 1/3 annotators in this group spent less than 30 seconds annotating an instance, while roughly 1/3 annotators spent more than 90 seconds per HIT.
It is hard to believe that annotators spending less than 30 seconds on a single instance can produce decent annotations.
This shows that the quality of the workforce is highly variable, making the quality of the annotation results unstable.

As another preliminary quality indicator, we compare how likely the annotations are rejected for not labeling any words as important.
Since we have stated clearly in the instructions that empty annotations will be directly rejected, those who submitted empty annotations can be seen as not even carefully reading the instructions.
We find that about 17\% of annotations of CoS-E from turkers with HIT approval rate larger than 80\% are empty annotations, indicating that they did not even try to fake that they worked on their job.
Contrarily, only about 1.7\% of annotators with HIT approval rates larger than 95\% submitted empty annotations.

Unfortunately, annotations that are not empty annotations are not guaranteed to be good annotations.
Since it is hard to evaluate the annotation results by any metrics, we do some in-browser checks for the rest annotations.
We find there exists obviously bad annotations among all three levels of HIT approval rates.
Those bad annotations include: Only labeling the first words in the instance, while the first word is often a stop word (commonly used word, e.g., is, are, the, a); failing to label some evidently important words in the instance; randomly labeling words as important, while most of the words selected are stop words or punctuation.
This indicate that even recruiting turkers with sufficiently high HIT approval rates cannot produce reasonable annotations.

We also explore the possibility of hiring turkers with master qualifications, but sadly there still exists some turkers that consistently produce low-quality annotations.
Obviously, they are just good at pretending that they did the job. 
Based on the above results, we find that the annotation quality can vary greatly using AMT, and even using qualifications provided by AMT does not guarantee reasonable results.
This highlights the importance of specifying the turker qualifications used in one's experiment.

The following experiments in the paper are conducted with the aids of "good turkers", in which we manually selected them based on some filtering rules, detailed in Appendix~\ref{app:amt}.

\section{How Do Task Instructions Affect the Annotated Results?}
\label{sec:instruction}
In this section, we turn our attention to how different instructions may affect the annotation results.
The goal of the annotation process is to identify parts in the text that the annotators think to be \textit{important}.
However, the concept of \textit{important} is quite vague and may vary from turkers to turkers, so the instructions are critical for guiding the annotators to provide sound annotations.

To understand the effect of task instructions on annotated results, we design several different instructions for the same set of datasets.
The instructions we use in this work are listed in Table~\ref{tab:instruction}. 
The design choices of these instructions will be specified in the following subsections.

\begin{table*}[ht]
    \centering
    \begin{tabular}{p{0.43\textwidth}|p{0.43\textwidth}}
    \hline
    
    SST2     &  CoS-E\\
    \hline
    \hline

    For each \textcolor{red}{\textbf{word}} in the review, if you think it to \textcolor{blue}{\textbf{support}} the reviewer's sentiment, please label them as 1; otherwise, label them with 0.     
    &
    For each \textcolor{red}{\textbf{word}} in the question, if you think it to \textcolor{blue}{\textbf{support}} the ground truth answer, please label them as 1; otherwise, label them with 0.\\
    \hline
    For each \textcolor{red}{\textbf{word}} in the review, if you think it is \textcolor{blue}{\textbf{critical for predicting}} the reviewer's sentiment, please label them as 1; otherwise, label them with 0. 
    &
    For each \textcolor{red}{\textbf{word}} in the question, if you think that it is \textcolor{blue}{\textbf{critical for predicting}} the ground truth answer, please label them as 1; otherwise, label them with 0. \\
    \hline
    Extract the \textcolor{red}{\textbf{parts}} in the review that you think them to \textcolor{blue}{\textbf{support}} the sentiment and label them as 1; otherwise, label them with 0.
    &
    Extract the \textcolor{red}{\textbf{parts}} in the question that you think them to \textcolor{blue}{\textbf{support}} the ground truth answer and label them as 1; otherwise, label them with 0.
    \\
    \hline
    Extract the \textcolor{red}{\textbf{parts}} in the review that you think them to be \textcolor{blue}{\textbf{critical for predicting}} the sentiment and label them as 1; otherwise, label them with 0.
    &
    Extract the \textcolor{red}{\textbf{parts}} in the question that you think to be \textcolor{blue}{\textbf{critical for predicting}} the ground truth answer and label them as 1; otherwise, label them with 0.
    \\
    \hline
    \hline
    
    For each \textcolor{red}{\textbf{word}} in the review, if you think that \textcolor{blue}{\textbf{removing it will decrease your confidence toward the reviewer's sentiment}}, please label them as 1; otherwise, label them with 0. 
    &
    For each \textcolor{red}{\textbf{word}} in the question, if you think that \textcolor{blue}{\textbf{removing it will decrease your confidence toward the ground truth answer}}, please label them as 1; otherwise, label them with 0.
    \\
    \hline
    
    \end{tabular}
    \caption{Instructions used in the human annotation for SST2 and CoS-E}
    \label{tab:instruction}
\end{table*}
\subsection{Instructions}
\subsubsection{Vague Instructions with Different Wordings}
\label{subsec:vague}
We first investigate how the annotation result will be like if we only present the turkers with vague instructions.
The instructions used are illustrated in the first block in Table~\ref{tab:instruction}.
We control two factors in the instructions:
The first factor is the targeting granularity that we expect the turkers to annotate.
By using the term "\textcolor{red}{word}", we expect the turkers to annotate the text based on the unit of a single word; the term "\textcolor{red}{part}" expect the turkers to label a continuous span in the text (phrase or constituent).
Example annotations that we hope to obtain are showcased in Table~\ref{tab:example}.
The second factor is how we describe "importance", which are highlighted in \textcolor{blue}{blue} in Table~\ref{tab:instruction}.
The phrases highlighted in \textcolor{blue}{blue} are used to described the annotation process in previous works listed in Section~\ref{subsec:previous}.
While those phrases do instruct the turkers to label important snippets in the text, those instructions are quite vague as we do not explicitly define what should be considered important and what should not.
These phrases may be similar in meaning, but we want to know whether they look differently for the turkers.
For both SST-2 and CoS-E, we construct four different tasks, using the instructions in the first block of Table~\ref{tab:instruction} respectively. 

\subsubsection{Vague Instructions with Example Annotations}
\label{subsec:vague+example}
We also explore what will the annotation results be like if we instruct the turkers with vague instructions but provide them with example annotations.
We want to know how well we can guide the turkers to follow the pattern in the example annotations.
We design two types of guidance, as illustrated in Table~\ref{tab:example}.
The upper part in Table~\ref{tab:example} attempts to guide the turkers to label the minimum words in the text whose removal will cause the turker unable to determine the ground truth answer; we call this kind of guidance \textbf{word guide}.
The lower part tries to tempt the turkers into labeling more complete and continuous span in the text, such as a complete noun phrase or verb phrase; we call this kind of guidance \textbf{span guide}.

For both SST-2 and CoS-E, we construct two different tasks, one with the \textbf{word guide} and one with the \textbf{span guide}. 
For each type of guidance, we provide four example annotations, one of them is illustrated in Table~\ref{tab:example}. 
The instructions we use is the "\textcolor{red}{parts}, \textcolor{blue}{support}" in the third row of Table~\ref{tab:instruction}.
\begin{table*}[ht]
    \centering
    \begin{tabular}{c|p{0.4\textwidth}p{0.4\textwidth}}
    \hline
    &SST-2     &  CoS-E\\
    \hline
    Word &
    one of the \colorbox{yellow}{best} films of the year with its exploration of the obstacles to happiness faced by five contemporary individuals ... a psychological \colorbox{yellow}{masterpiece} .
    &
    If you're attending \colorbox{yellow}{school} and are \colorbox{yellow}{falling asleep} you're likely experiencing what? (boredom) (malaria, graduate, inspiration, detention)   
    \\
    \hline
    Span &
    \colorbox{yellow}{one of the best films of the year} with its exploration of the obstacles to happiness faced by five contemporary individuals ... \colorbox{yellow}{a psychological masterpiece} .
    &
    If you're \colorbox{yellow}{attending school and are} \colorbox{yellow}{falling asleep} you're likely experiencing what? (boredom) (malaria, graduate, inspiration, detention)
    \\
    \hline
    \end{tabular}
    \caption{One of the example annotations we give the turkers in SST-2 and CoS-E.
    Example annotations in the upper row guide the turkers to label only the words most related to the ground truth, while example annotations in the lower row aim to lead the annotator to extract longer and complete spans in the text.
    The parentheses in SST-2 are the ground truth sentiment.
    The 1st parenthesis in CoS-E is the ground truth answer, and the 2nd contains the other four wrong answers separated by a comma.}
    \label{tab:example}
\end{table*}

\subsubsection{Precise Instructions}
Last, we try to give the turkers some concrete operational definition of "important words": a word should be considered important if removing it will decrease one's confidence toward the ground truth answer.
Thus, a turker needs to anticipate the counterfactual outcome of the removal of each word to determine the importance of each word.
We call this kind of instruction \textbf{word remove}.
We expect the turkers to give annotations similar to the example annotations of the \textbf{word guide}.
\label{subsec:precise}

\subsection{Annotation Results}
\begin{table}[h]
    \centering
    \begin{tabular}{|c|c|c|c|c|}
    \hline
        Instructions & \multicolumn{2}{c}{SST-2} & \multicolumn{2}{c}{CoS-E}  \\
        \cline{2-5}
         & \(l_{avg}\) & \(s_{avg}\)  & \(l_{avg}\) & \(s_{avg}\) \\
         \hline
         word, support & 0.36 & 0.21 & 0.32 & 0.17
         \\
         word, critical & 0.29 & 0.17 & 0.43 & 0.24 
         \\
         part, support & 0.32 & 0.24 & 0.30 & 0.17
         \\ 
         part, critical & 0.38 & 0.27 & 0.36 & 0.19
         \\
         \hline \hline
         word, remove & 0.21 & 0.10 & 0.20 & 0.10
         \\
         \hline \hline
         word guide & 0.36 & 0.24 & 0.33 & 0.16
         \\ 
         span guide &  0.45 & 0.28 & 0.38 & 0.24
         \\
         \hline
    \end{tabular}
    \caption{The annotation statistics of SST-2 and CoS-E under different instructions.}
    \label{tab:vague statistics}
\end{table}

The statistics of the annotations from different instructions are shown in Table~\ref{tab:vague statistics}.
We report the average annotation length \(l_{avg}\),
\begin{equation}
    l_{avg} = \frac{1}{N}\sum_{j = 1}^{N} \frac{i_j}{l_j},
\end{equation}
where \(N\) is the total instances annotated by all the turkers, \(i_j\) is the number of words labeled important in instance \(j\), and \(l_j\) is the number of words in instance \(j\).
We also report the average percent of stop words per annotation, \(s_{avg}\),
\begin{equation}
    s_{avg} = \frac{1}{N}\sum_{j = 1}^{N} \frac{s_j}{i_j},
\end{equation}
where \(s_j\) are the number of stop words that are labeled as important in instance \(j\).

Comparing the average annotation length, we do not observe that the annotations instructed with \textcolor{red}{part} to be longer than those instructed \textcolor{red}{word}.
This may be largely due to the two terms seeming to refer to the same thing for the turkers when we do not explicitly define their difference.
We also fail to uncover any meaningful correlations between those vague instructions and their corresponding statistics.
However, when provided with concrete example annotations, the \textbf{span guide} does give a significantly longer average annotation length compared to the annotation length of \textbf{word guide}, indicating that the turkers are successfully guided to provide longer annotations.
It can also be observed that the average annotation length of \textbf{word remove} is the shortest among all instructions types for both SST-2 and CoS-E, implying that giving precise instructions is the best way to guide the turkers to give annotations that only highlights the minimal number of important words.

Examining the results of the average ratio of stop words per annotation, we find them to follow our expectation: \textbf{word guide} does induce the turkers to label fewer stop words in their annotations compared to \textbf{span guide}. 
Also, the \textbf{word remove} results in the least stop words labeled, since one won't expect the removal of stop words to alter the answer.

To further compare whether the annotations obtained under the same instructions are more similar or not, we compare the Cohen $\kappa$\citep{cohen1960coefficient} between annotations from different annotators.
Cohen $\kappa$ is a metric for measuring the pair-wise agreement between two annotators.
Fixing an instruction, for each instance in the dataset, we have 3 different annotations from 3 turkers.
Thus, we can calculate the \textit{intra-instruction} agreement of an instance as the average of pair-wise Cohen $\kappa$ among two of the three annotators.
The average intra-instruction agreement of instruction is obtained by averaging the intra-instruction agreement of each instance in the dataset.
When comparing the agreement between different instructions, we compute the \textit{cross-instruction} pair-wise agreements of the annotations from different instructions, resulting in a total of $3\times3=9$ Cohen $\kappa$s.
We average the 9 Cohen $\kappa$s as the cross-instruction agreement score of an instance and average the per-instance cross-instruction agreements as the agreement score between two instructions.

\begin{figure}[t!]
\centering
\includegraphics[clip, trim =  0 0 0 40, width=0.76\linewidth]{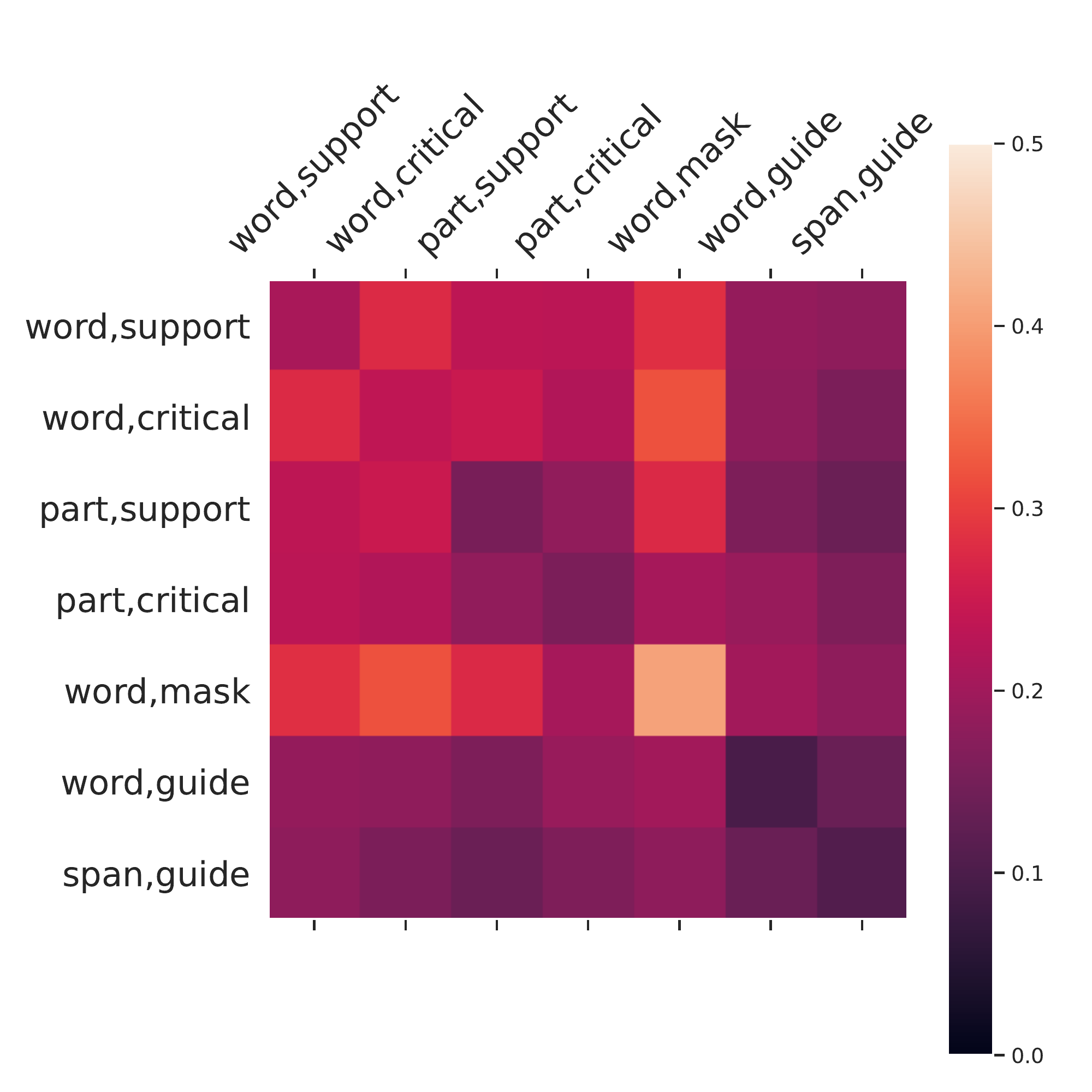}
\caption{The average pair-wise agreement scores between annotations from different or same instructions.}
\label{fig:sst2 kappa}
\end{figure}

The agreement scores between annotation between all instructions of SST-2 are shown in Figure~\ref{fig:sst2 kappa}.
First, it is easy to observe that the highest agreement score is the intra-instruction score of word remove, showing that giving precise instructions will lead to better agreements between annotations.
On the other hand, the intra-instruction agreement scores for other instructions are much lower, indicating that vague instructions will result in high variability among annotations, and this issue cannot be eased by providing the annotators with example annotations.

Next, we turn our attention to the cross-instruction agreement scores, and we observe that most of them have a poor agreement.
Quite surprising is the cross-instruction agreement scores between \textbf{word/span guide} and all other instructions are quite low, suggesting that the turkers' annotation with and without example annotations can be quite different.
We also observe that annotations obtained from \textbf{word remove} have fair agreement scores with most other instructions on average, this is possibly due to the fact that the important words labeled by \textbf{word remove} tend to be the subset of the important words labeled under other instructions. 

The agreement scores of CoS-E are presented in Figure~\ref{fig:cose kappa}.
We discuss the results of CoS-E in Appendix~\ref{app:cose kappa}

Overall, we find that using different instructions will result in quite different annotation results.
We also show that one can use the instructions to guide the turkers to provide annotations that display certain patterns.
This again highlights the importance of providing full instructions used in the annotation process, such that those who use the annotation data can fully understand if there exist any systematical biases in the annotations.

\begin{table*}[t!]
    \centering
    \begin{tabular}{c|ccccccc}
        Method & \makecell{word \\support} & \makecell{word\\ critical} & \makecell{part\\ support} & \makecell{part \\critical} & \makecell{word \\remove} & \makecell{word\\ guide} & \makecell{span\\ guide} \\
        \hline
        Gradient & 0.42&0.42&0.39&0.42&0.42&0.42&0.41 \\
        Integrated Gradient & 0.39&0.40&0.42&0.35&0.47&0.37&0.34 \\
        Smooth Gradient &  0.41&0.43&0.40&0.42&0.42&0.40&0.41 \\
        Occlusion & 0.32&0.32&0.34&0.28&0.34&0.30&0.26\\
        Lime & 0.30&0.32&0.32&0.31&0.33&0.29&0.27 \\
        
    \end{tabular}
    \caption{Average maximum agreement score between the explanations from different methods and annotations obtained under different instructions for SST-2.}
    \label{tab:explanation method}
\end{table*}

\section{How Do the Annotations from Different Instructions Affect the Plausibility of Explanation Methods?}

When evaluating the plausibility of explanation methods, human annotations obtained from crowd-sourcing are used as the ground truth explanations.
In this section, we aim to know whether the human annotations obtained from different instructions will affect the plausibility of an explanation method.
We evaluate the plausibility of five input feature-based explanation methods with seven kinds of annotations obtained using different instructions from Section~\ref{sec:instruction}.

\subsection{Experiment Setting}
Given a trained model and a piece of text, an explanation method based on the input features will score the importance of each word using a scalar; the larger the score is, the more important the word is.

In our experiments, we adopt five commonly used input feature-based explanation methods: Gradient, Integrated Gradient, Smooth Gradient, Occlusion, and LIME.
The first three methods are the gradient-based explanations.
The Gradient method calculates the derivative of the ground truth's logit (of the model) with respect to the word embedding of a word in the input text, and uses the $l_2$-norm of the derivative as the word importance score. 
Smooth Gradient~\cite{smilkov2017smoothgrad} samples multiple noised word embeddings of the target word with Gaussian noises, using those noised embeddings to calculate the corresponding derivatives as in the Gradient method, and takes the average of the norm of the derivatives as the importance score of the target word.
Integrated Gradient~\cite{sundararajan2017axiomatic} calculates the integral of the gradient along the path from a baseline to the target word, and uses the norm of the integral as the importance score of the target word.
The Occlusion method assigns the importance score as the change in the ground truth's logit before and after the target word is removed.
LIME~\cite{ribeiro2016should} fits a linear model on the local perturbations of the input text and their corresponding logit, and uses the $l_1$-norm of the linear model's coefficients as the importance score of each word.

We fine-tune SST-2 and CoS-E on BERT-base-uncased as our model.
We then use the five input-based explanation methods to generate the importance score for the 100 instances that we asked the turkers to label.
Since the importance scores from the explanation methods are positive real numbers, we need to convert them to 0-1 labeling so we can compute the plausibility against the annotations from the turkers.
We use the following procedure:
Fixing an instruction and an explanation method, for each instance, we will have three annotation results and one explanation from the explanation method.
We take the top-$k$ important word from the explanation method and label them as important, where $k$ is the mean annotation lengths of the three human annotations and taking its floor.
After converting the explanation method's importance score to 0-1 score, we compute the Cohen $\kappa$ with the three human annotations and take the maximum $\kappa$ as the agreement score for this instance.
By averaging the maximum agreement score from each instance, we obtain the agreement score between an explanation method and the annotation from an instruction.
This is the plausibility of the explanation method when using the instruction.

\subsection{Results}
The results for SST-2 are shown in Table~\ref{tab:explanation method}.
Comparing different rows, we can observe that explanations obtained from gradient-based explanation methods, in general, have better agreement with all the turkers annotations.
For Gradient and Smooth Gradient, the agreement scores are quite invariant when using different instructions.
Contrarily, Integrated Gradient shows observable instability for different instructions.
Another interesting point to note is the best agreement for Occlusion happens when the instruction is \textbf{word remove}; this should be expected since the way Occlusion calculates the word importance is quite similar to how the turkers are instructed to label the word importance.
Despite this similarity, the agreement score between Occlusion and \textbf{word remove} is still much lower than agreement scores between gradient-based explanations and \textbf{word remove}.
Overall, the results show that when evaluating the plausibility of an explanation method, how the annotations are obtained, in terms of their instructions used during the annotation process, does matter.

\section{Discussion}
In this work, we first point out an important problem in the benchmarks for Interpretable NLP: the details for the crowd-sourced annotation process are largely missed.
To show that the missing details are decisive to the human annotation results, we conduct substantially controlled experiments using AMT.
Our results reveal that the qualification of the turkers will significantly influence the quality of the annotations.
We also illustrate that the turkers will provide different annotations when given different instructions, and the turkers can be successfully guided to provide annotations with the desired pattern using example annotations or precise instructions.
Last, we reveal that the plausibility of an explanation method can be quite different when evaluated with annotations acquired by different instructions.

As a pioneering work, we hope to raise the attention on how the annotations for Interpretable NLP are obtained.
While providing the details of crowd-sourced annotation doesn't guarantee others to reproduce similar annotation results, this is the minimum requirement for others to understand how the annotations are obtained.
When the users try to evaluate their proposed method on the annotated datasets, only if they are fully aware of the annotation process' detail can they understand what their experiment results are built on.
When one is not well-informed of how the dataset is annotated, hardly can one know if there exists any systematical bias during the annotation process and making the annotations prefer certain explanation methods.
Any conclusions drawn upon those datasets should be examined carefully.
We thus call for a more rigorous examination of how we report the details in the crowd-sourced annotation process. 

\section*{Ethical Statement}
In this work, we recruit turkers to conduct human evaluation.
To ensure that the turkers will not be exposed to offensive contents or data with social biased, the authors carefully scrutinize the SST-2 and CoS-E dataset and select 100 data instance in each of the two datasets.

It is also important to make sure that the human annotators are reasonably paid for their labor. 
The turkers are paid USD 0.08 for each HIT.
Since we expect each HIT can be completed in 30 to 40 seconds, the turkers will be paid USD 7.2 to 9.6 per hour.
Considering the turkers are mostly not from North America and Europe, we believe that the payment is very reasonable.

We believe that our work do not violate the ethical guidelines of AAAI 2022.

\section*{Acknowledgement}
We want to thank the anonymous reviewers for providing insightful and actionable suggestions on our work. 
We thank National Center for High-performance Computing (NCHC) of National Applied Research Laboratories (NARLabs) in Taiwan for providing computational and storage resources.

\bibliography{aaai22}

\appendix

\section{Eraser Benchmark}
Eraser benchmark~\citep{deyoung2020eraser} comprises multiple datasets and tasks for which human annotations of "rationales" ("supporting evidence") have been collected.
We summarize the details of human annotation in Table~\ref{tab:annotation details}.
For Movie Reviews, Evidence Inference, and BoolQ, the annotation process we refer to is the annotation collected by ERASER.
For the rest four datasets, the annotation processes we refer to are those reported in their original paper (instead of ERASER).
We are not sure whether FEVER is collected using a crowd-source platform since there are few details on how the annotators are selected.

In Common Sense Explanation~\citep{rajani2019explain}, turkers are prompted with the following question: "Why is the predicted output the most appropriate answer?"
They are instructed to highlight relevant words in the question that justify the ground truth answer and provide an open-ended explanation based on the highlighted words.
In e-SNLI~\citep{camburu2018snli}, turkers are also asked to highlight important words in the text and form natural language explanations using those highlighted words.
For MultiRC~\citep{khashabi2018looking} and FEVER~\citep{thorne2018fever}, the annotation processes are somewhat complicated, and we refer the readers to the original paper.
\begin{table*}[ht]
    \centering
    \begin{tabular}{|p{0.14\linewidth}|cp{0.3\linewidth}cc|}
    \hline
        \centering Dataset & Crowd-source & Worker Qualification & Payment & Detailed Instructions \\
        \hline \hline
        \centering Movie Reviews  & Upwork & 2 Fluent English speakers & \cmark & \xmark \\
        \hline
        \centering Evidence Inference & Upwork & 4 medical professional fluent in English who have passed a pre-test & \cmark & \xmark \\
        \hline
        \centering BoolQ & AMT & HIT approval rate $>98\%$, total approved HITs $>10K$, live in AU, NZ, CA, US, GB & \cmark & \xmark \\
        \hline
        \centering Common Sense Explanation & AMT & \centering\xmark & \xmark & \cmark \\
        \hline
        \centering e-SNLI & AMT & \centering\xmark & \xmark & \cmark \\
        \hline
        \centering MultiRC & AMT & \centering\xmark & \xmark & \cmark \\
        \hline 
        \centering FEVER & \xmark  & \centering\xmark & \xmark & \cmark \\
        \hline
        
    \end{tabular}
    \caption{Annotation details for the seven datasets in ERASER.}
    \label{tab:annotation details}
\end{table*}

\section{Input Methods}
\label{sec:input methods}
In this section, we will use the term "saliency" to refer to "word/token importance".
Also note that the input methods can only compute the sub-word token's saliency; in order to convert the sub-word token saliency to the word saliency, we use the following procedures:
If a word is tokenized into "A", "B" two tokens, and the input method scores the token saliency by $s_A, s_B$, then we obtain the word saliency by $max(s_A,s_B)$.
\subsection{Gradient}
When we use the gradient to compute the token saliency of an input sentence for explaining a trained model, we first convert the sentence into its corresponding input embeddings by looking up the embedding table of the model.
Then we forward the input embeddings to obtain the logit scores, which is the final prediction of the model.
There will be $n$ logit scores for a $n$-way classifier.
In our case of SST-2, $n=2$; in our case of CoS-E, $n$ will be the number of tokens in the question (refer to Appendix~\ref{app:finetune} for more details.).
We take the derivative of the logit score of the ground truth label's class against the input embeddings. 
This will give as the gradient input embedding., which has the same shape as the input embeddings.
For each token's gradient, we take the $l_2$ norm of the vector as the token saliency.
\subsection{Smooth Gradient}
In smooth gradient, instead of forwarding the model with the original input embeddings, we forward the model with $K$ "noised" input embeddings.
Each of the $K$ input embeddings is obtained by adding a Gaussian noise on the original input embeddings.
Then we calculate the gradient of the $K$ noised input embeddings as in the Gradient method.
Since for each token, we will have $K$ gradient norms, we average those gradient norms as the token saliency scored by the smooth gradient.
\subsection{Integrated Gradient}
When using the integrated gradient to calculate the token saliency by integrating the gradient over the path along a baseline vector to the target vector.
The target vector is the input embeddings, and we select the zero vector as the baseline vector.
It is impossible to calculate the integral, so we segment the path into multiple segments, and sum the gradient of each segment as the surrogate of the path integral.
We also take the $l_2$ norm of the surrogate of the path integral to obtain the final saliency of each token.
\subsection{Occlusion}
In occlusion, each token's saliency is calculated by subtracting the ground truth logit when the token is presented and when the token is not presented.
Since we use BERT as our base model, we can use the [MASK] token to substitute the token of interest and forward the masked sentence to the model to obtain the ground truth logit when the token is not presented.
\subsection{LIME}
In LIME, we train a linear classifier on the local perturbation of the original input sentence.
The perturbation is sampled as follows: 
First, we determine how many words in the input sentence should be removed.
This is done by uniformly selecting a number ranging from 0 to the length of the sequence.
Then, each word in the sentence has an equal probability to be removed. 
After we removed the words from the input sentence, we obtain the perturbed version of the input sentences.
The perturbed sentences are forwarded to the trained classifiers to obtain the ground truth logit score.
We then use the perturbed sentences' bag of word representations and their corresponding logit scores to fit a linear regression model.
The regression model will give each token in the sentence a weight, which is the weight of the regression model.
We take the $l_1$ norm of the weight to represent the token importance.

\section{AMT Details}
\label{app:amt}
Since using bad turkers may make our later experiments inconclusive, we manually select high-quality turkers based on the following guidelines:
We first publish 10 tasks that are related to our annotation task but not exactly the same, and recruit turkers with HIT approval rates greater than 95\% and have th number of HITs approved greater than 5K.
Based on their annotation results, we manually select annotators that spend a reasonable duration on each HIT and persistently produce valid annotations.
During the selection, we stay as open-minded as we can; we do not only select turkers that annotate following a pre-defined pattern. 
The experiments of Section~\ref{sec:instruction} are completed by those turkers that pass our filtering.
We pay the turkers for USD 0.08 for each HIT.
\begin{figure}[t!]
\centering
\includegraphics[clip, trim =  0 0 0 0, width=0.99\linewidth]{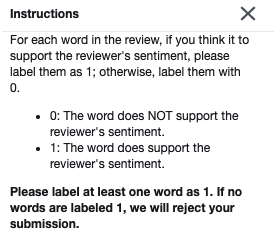}
\caption{An example instruction for the turkers for SST-2.}
\label{fig:instruction}
\end{figure}

\begin{figure}[t!]
\centering
\includegraphics[clip, trim =  0 0 0 0, width=0.99\linewidth]{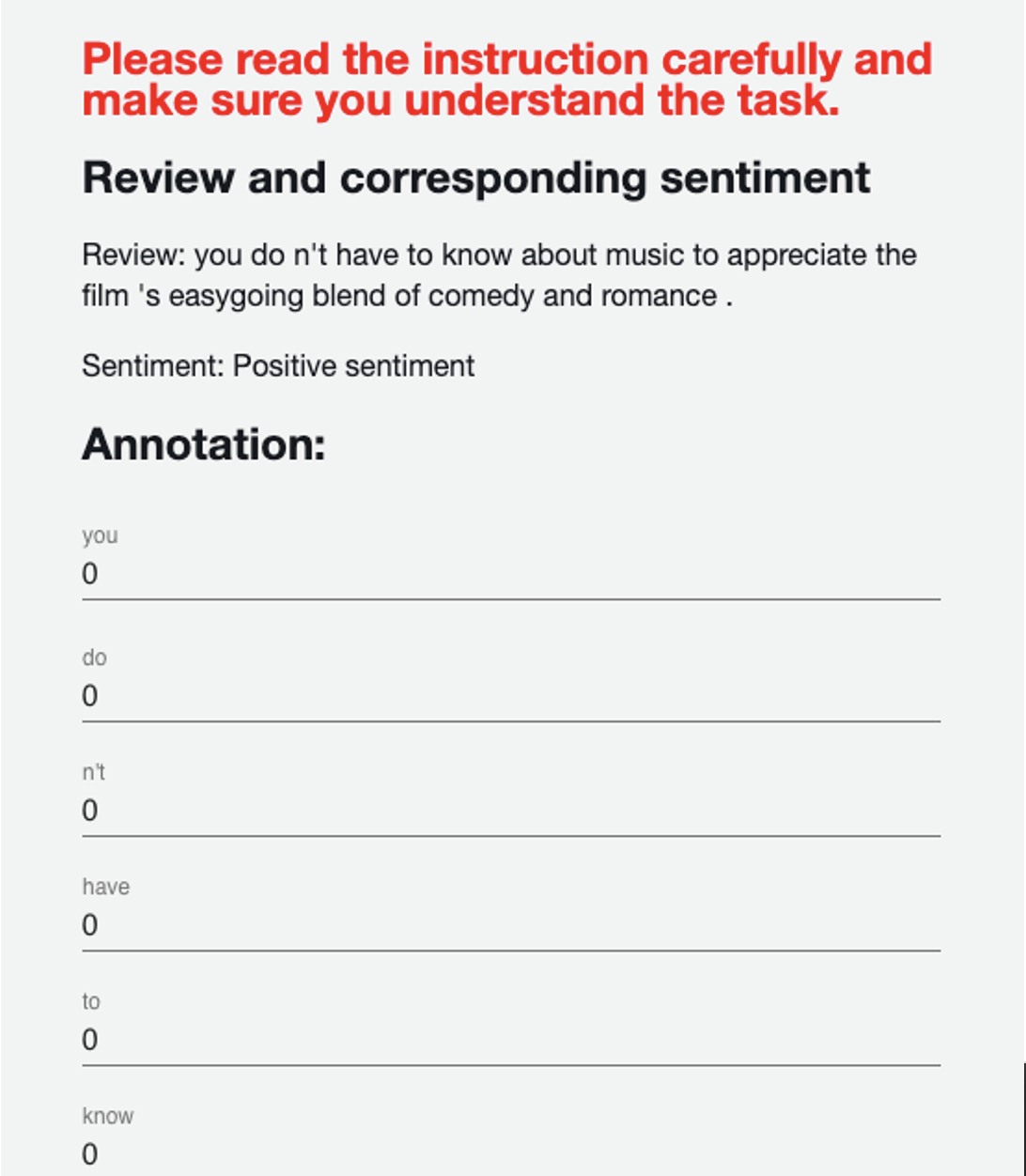}
\caption{The annotation interface used in this work. 
The interface for SST-2 and CoS-E are the same.
Here we only show the interface of SST-2.
}
\label{fig:interface}
\end{figure}

\section{CoS-E Experiment Results}
\subsection{Agreemnent}
\begin{figure}[t!]
\centering
\includegraphics[clip, trim =  0 0 0 0, width=0.99\linewidth]{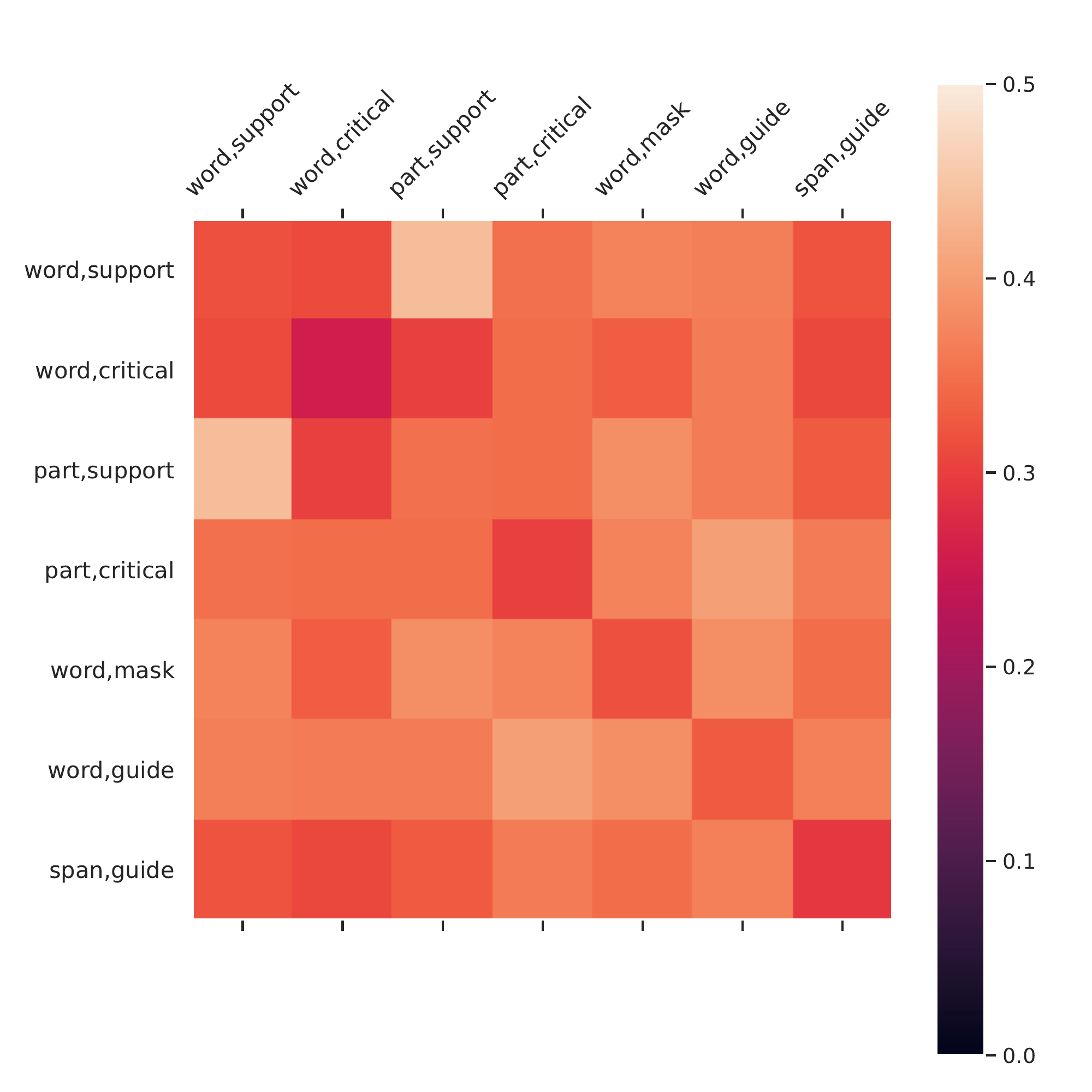}
\caption{The average pair-wise agreement scores between annotations from different or same instructions of CoS-E.}
\label{fig:cose kappa}
\end{figure}

The result of the agreement of CoS-E is shown in Figure!\ref{fig:cose kappa}.
We find that most of the results observed in SST-2 do not hold. 
This may arise from two reasons: 
First, the sentences in CoS-E are shorter than those in SST-2 on average, which possibly makes them easier for the annotators to focus on similar portions in the text. 
This makes the Cohen $\kappa$ quite high in Figure~\ref{fig:cose kappa}.
Second, the nature of CoS-E makes the turkers easier to agree on their annotations. 
Since the dataset tests one's commonsense, we can expect that commonsense should be more similar among different turkers, and thus making the Cohen $\kappa$s higher than those of SST-2.
On the other hand, what a turker thinks to make a good/bad movie can be quite subjective, and making the Cohen $\kappa$s lower between annotations.

This shows that the dataset's nature plays an important role in the relationship between annotation results and instructions.
\label{app:cose kappa}

\subsection{Human Annotations vs Explanation Methods}
We find that the explanation methods and human annotation's agreement are not quite like those of SST-2.
This may stem the following reason:
The model fine-tuned on CoS-E do not have very high performance on the development set, which may make the input methods unable to explain the model's decision well.

This show the difficulty when one tries to compare the input methods' explanation with human's annotations.
It also reveals that the results one obtain in Interpretable AI on one dataset will be highly likely not able to generalize on the other dataset, especially when human annotations are involved.

\begin{table*}[h]
    \centering
    \begin{tabular}{c|ccccccc}
        Method & \makecell{word \\support} & \makecell{word\\ critical} & \makecell{part\\ support} & \makecell{part \\critical} & \makecell{word \\remove} & \makecell{word\\ guide} & \makecell{span\\ guide} \\
        \hline
        Gradient &0.58&0.61&0.59&0.57&0.61&0.56&0.55 \\
        Integrated Gradient & 0.32&0.36&0.32&0.32&0.30&0.29&0.29 \\
        Smooth Gradient & 0.60&0.61&0.60&0.57&0.60&0.57&0.55 \\
        Occlusion&0.30&0.31&0.32&0.31&0.29&0.30&0.30\\
        Lime & 0.25&0.21&0.22&0.21&0.18&0.22&0.22 \\
        
    \end{tabular}
    \caption{Caption}
    \label{tab:cose model}
\end{table*}
\section{Fine-tune}
\label{app:finetune}
\subsection{SST-2}
We fine-tune SST-2 using the same procedure of ~\cite{devlin2019bert}. The hyperparameters are listed in Table~\ref{tab:hpp}. 
We do not perform early stopping. 
The accuracy on the development set after training is 91.2\%. 
\subsection{CoS-E}
For CoS-E, we first convert the question and 5 options into the following form: "[CLS] Question [SEP] Option 1, Option 2, Option 3, Option 4, Option 5 [SEP]".
We add a classifier head on top of the pre-trained BERT model. 
The model is trained as an extractive QA model: the first classifier needs to point out the start token of the answer, and the second classifier is trained to point out the end token.
In our setting, it is only necessary to point out the start token, since the end token is the previous token of the next comma after the start token.
Our fine-tuning procedure of CoS-E is quite different from the canonical way for fine-tuning multiple-choice datasets using BERT. 
We choose to fine-tune the model this way because we need to compute the derivative of the ground truth answer's logit with respect to the input, and this can not be done when we use the canonical way for fine-tuning multiple-choice dataset.
The development set accuracy (the probability that the start token classifier points to the first token of the ground truth answer) after training is 48.5\%. 
While this accuracy is not high, it is still far higher than random guessing. 
Both of the datasets are fine-tuned using the Huggingface based on Pytorch. 
\begin{table}[t]
    \centering
    \begin{tabular}{|c|c|c|}
    \hline
         & SST-2 & CoS-E \\
    \hline
    Learning rate     & 5e-5 & 3e-5\\
    Warm-up steps &100 &100\\
    Random seed & 21616 & 21616\\ 
    Batch size & 128& 128\\
    Training Epochs &5 &5 \\
    \hline
    \end{tabular}
    \caption{Hyperparameters used to fine-tune SST-2 and CoS-E.}
    \label{tab:hpp}
\end{table}
\end{document}